\documentclass[conference]{IEEEtran}
\IEEEoverridecommandlockouts
\usepackage{cite}
\usepackage[numbers,sort&compress]{natbib}

\usepackage[utf8]{inputenc}
\usepackage{amsmath,amssymb}
\usepackage{graphicx}
\usepackage{bm}
\usepackage{enumerate}
\usepackage{comment}
\usepackage{xcolor}
\usepackage{url}
\usepackage{color,soul}
\usepackage{subcaption}
\usepackage{float}

\definecolor{light-gray}{gray}{0.8}

\usepackage{amsmath,amssymb,amsfonts}
\usepackage{algorithmic}
\usepackage{graphicx}
\usepackage{textcomp}
\usepackage{xcolor}
\usepackage{subcaption}

\def\BibTeX{{\rm B\kern-.05em{\sc i\kern-.025em b}\kern-.08em
    T\kern-.1667em\lower.7ex\hbox{E}\kern-.125emX}}

\makeatletter
\newcommand{\linebreakand}{%
  \end{@IEEEauthorhalign}
  \hfill\mbox{}\par
  \mbox{}\hfill\begin{@IEEEauthorhalign}
}
\makeatother

\newcommand{\etal}{\textit{et al}. }

\begin{document}

\title{Regional Style and Color Transfer\\}

\author{

\small 

\begin{tabular}[t]{c@{\extracolsep{8em}}c} 

1\textsuperscript{st} Zhicheng Ding  & 2\textsuperscript{nd} Panfeng Li \\
\textit{Fu Foundation School of Engineering and Applied Science} & \textit{Department of Electrical and Computer Engineering} \\
\textit{Columbia University} & \textit{University of Michigan} \\
New York, USA & Ann Arbor, USA \\
zhicheng.ding@columbia.edu & pfli@umich.edu \\

\\

3\textsuperscript{rd} Qikai Yang & 4\textsuperscript{th} Siyang Li \\
\textit{Department of Computer Science} & \textit{Lubin School of Business} \\
\textit{University of Illinois Urbana-Champaign} & \textit{Pace University} \\
Urbana, USA & New York, USA \\
qikaiy2@illinois.edu & lisiyang98@hotmail.com \\

\\

\multicolumn{2}{c}{5\textsuperscript{th} Qingtian Gong}  \\
\multicolumn{2}{c}{\textit{Fu Foundation School of Engineering and Applied Science}} \\
\multicolumn{2}{c}{\textit{Columbia University}}\\
\multicolumn{2}{c}{New York, USA} \\
\multicolumn{2}{c}{qg2159@columbia.edu} \\

\end{tabular}
}

\maketitle

\begin{abstract}
This paper presents a novel contribution to the field of regional style transfer. Existing methods often suffer from the drawback of applying style homogeneously across the entire image, leading to stylistic inconsistencies or foreground object twisted when applied to images with foreground elements such as person figures. To address this limitation, we propose a new approach that leverages a segmentation network to precisely isolate foreground objects within the input image. Subsequently, style transfer is applied exclusively to the background region. The isolated foreground objects are then carefully reintegrated into the style-transferred background. To enhance the visual coherence between foreground and background, a color transfer step is employed on the foreground elements prior to their reincorporation. Finally, we utilize feathering techniques to achieve a seamless amalgamation of foreground and background, resulting in a visually unified and aesthetically pleasing final composition. Extensive evaluations demonstrate that our proposed approach yields significantly more natural stylistic transformations compared to conventional methods.


\end{abstract}


\begin{figure*}
  \centering
  \includegraphics[width=\linewidth]{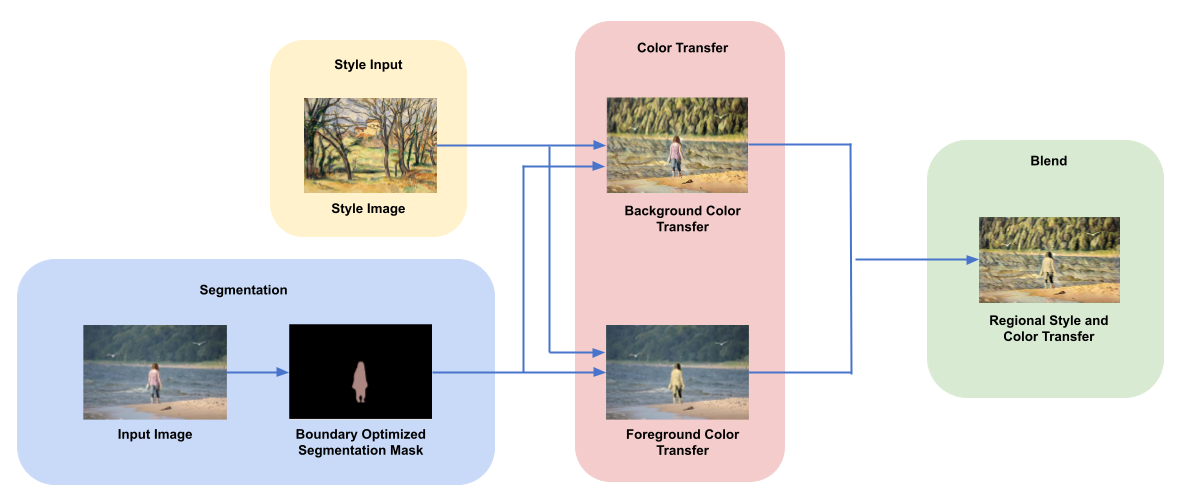}
  \caption{Regional style and color transfer framework}
  \label{framework}
\end{figure*}

\section{Introduction}
In the domain of computer vision, recent advancements in image segmentation techniques~\cite{li2024cpseg} have had a profound impact. These advancements facilitate the precise delineation of objects and regions within images. Traditional segmentation methods primarily relied on handcrafted features and heuristic algorithms. However, the emergence of deep learning-based approaches, particularly convolutional neural networks (CNNs) \cite{peng2023autorep, chang2023explainable}, has revolutionized this field. CNN-based models like U-Net and Mask R-CNN~\cite{xin2024parameter} have demonstrated exceptional performance in segmenting objects with intricate shapes and diverse appearances. This paves the way for more accurate and efficient image understanding tasks.

In parallel, The field of style transfer has concurrently witnessed significant progress in recent years, fueled by the growing interest in neural style transfer algorithms. These algorithms aim to apply the artistic style of a reference image to a target image while faithfully preserving content of the target image. Early methods employed optimization-based approaches. However, with the advent of deep learning, neural style transfer has evolved to leverage pre-trained CNNs like VGG and ResNet~\cite{yao2023ndc, deng2024multi}, enabling the achievement of impressive results in real-time style transfer applications. These advancements have unveiled new avenues for creative expression and artistic rendering within the realm of digital media.

Similarly, color transfer techniques have also undergone remarkable developments, particularly within the context of image enhancement and color harmonization. These techniques aim to transfer the color distribution of a source image to a target image while maintaining its content and structural integrity. Traditional color transfer algorithms often relied on histogram matching and statistical methods. However, modern approaches leverage deep learning architectures to learn intricate mappings between color distributions~\cite{Reinhard}. These deep learning-based color transfer models offer superior performance and flexibility, enabling the creation of more natural and aesthetically pleasing color adjustments in images.

To address the limitations associated with unnatural style transfer, this paper presents a novel regional artistic style and color transfer approach. We first employ object segmentation to process the image and generate a mask. This mask facilitates the division of the foreground object (e.g., person) and the background (e.g., scenic view, road, room). Subsequently, color transfer and style transfer are applied to the foreground and background regions, respectively. Finally, the two stylistically altered images are seamlessly blended together.

In summary, our paper makes the following key contributions:
\begin{itemize}
    \item We propose a regional style and color transfer approach specifically designed to overcome the unnatural effects associated with global style or color transfer applied to the entire image.
    \item Through extensive experimentation, we demonstrate the advantages of applying style and color transfer within distinct image regions, resulting in a more natural aesthetic outcome.
    \item Our method possesses the inherent capability to be extended to establish a correspondence between painted images and video frames, thereby enabling the creation of stylized films.
    \item Leveraging effective segmentation techniques, our approach offers a concise and effective framework for seamlessly applying and switching between style transfer algorithms.
\end{itemize}

\section{Methods}
This section details the comprehensive approach adopted in this study to elucidate the influence of regionally applied stylistic and chromatic alterations. The proposed framework at fig \ref{framework} takes two input images: a content image and a style image. The model strategically partitions the style transfer process, applying the stylistic elements to the background region and the color transfer to the foreground region. To achieve this regional manipulation, semantic segmentation~\cite{deeplabv3plus2018} is leveraged to delineate the person foreground. Subsequently, the style-transferred background and color-transferred foreground are seamlessly integrated using a blending technique, effectively mitigating any potential artifacts at the boundary.

\begin{figure}[h!]
  \centering
  \begin{subfigure}[T]{0.4\linewidth}
    \includegraphics[width=\linewidth]{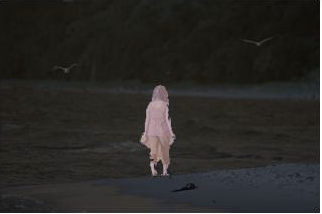}
    \caption{No Boundary Optimization}
  \end{subfigure}
  \begin{subfigure}[T]{0.4\linewidth}
    \includegraphics[width=\linewidth]{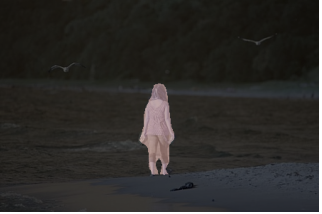}
    \caption{With Boundary Optimization}
  \end{subfigure}
  \caption{Comparison of Segmentation Boundary with or without Optimization}
  \label{segmentation_mask}
\end{figure}

\subsection{Segmentation and Boundary Optimization}
\label{sec:segmentation}

\noindent\textbf{Semantic Segmentation} Our image segmentation approach leverages the DeepLabv3+ \cite{deeplabv3plus2018}. This model incorporates Atrous Spatial Pyramid Pooling (ASPP) \cite{zhao2017pyramid} alongside an encoder-decoder structure to capture rich contextual information. ASPP employs a series of atrous convolutions with varying dilation rates, enabling the model to learn features at diverse scales. The encoder progressively encodes these features across scales, while the decoder refines them to generate high-resolution segmentation masks. We trained the model on the PASCAL VOC 2012 dataset \cite{pascal-voc-2012}, encompassing 20 object classes including "person". This dataset comprises 11,530 images with 27,450 annotated objects and 6,929 segmentation. A ResNet-101 backbone was utilized with a stride of 16 to expedite the training process. This configuration achieved a segmentation performance of 82.1\% mean Intersection-over-Union (mIoU) on the PASCAL VOC 2012 validation set. Notably, with 64 channels, the model attained an mIoU of 78.21\%.

\noindent\textbf{Boundary Optimization} Person object boundaries extracted from segmentation masks often exhibit noise. To facilitate independent style and color application to distinct regions, precise boundaries are crucial. We address this by employing a two-step boundary optimization process. Initially, Canny edge detection is applied to the original image. Subsequently, a Breadth-First Search (BFS) algorithm traverses each point on the binary person mask, starting from the periphery and moving inwards. For each point, 1) If the closest edge point resides within the mask, the point is removed, and its unvisited neighbors are added to a First-In-First-Out (FIFO) queue. 2) If the nearest edge point lies outside the mask, the point is skipped, and the traversal continues. Finally, Euclidean distance-based smoothing is applied to refine the extracted boundary.

\subsection{Artistic Style Transfer}

We followed the idea of fast style transfer \cite{engstrom2016faststyletransfer} 
to train the style transfer. This method reduces the total amount of parameters in the model to achieve faster style transfers with pre-trained style. Given a layer $l$ of the model, $N_l$ is the number of feature maps, and  $M_l$ is the size of each feature map. The output of a layer can be stored in a matrix $F^l \in R^{N_l \times M_l}$  where $F_{ij}^l$ is the activation of the $i^{th}$ filter at position $j$ in layer $l$. The Gram Matrix from style representation is expressed: $G_{ij}^{l} = \sum_k{F_{ik}^l F_{jk}^l}$.

The squared-error loss between the original image and the generated images is defined as following: 
\begin{equation}
L_{content} = \frac{1}{2}\sum_{i,j}(F_{ij}^l - P_{ij}^l)^2
\end{equation}
\begin{equation}
L_{style} = \sum_{l=0}^{L} \frac{w_l}{4 N_l^2 M_l^2}
\sum_{i,j}(G_{ij}^l - A_{ij}^l)^2
\end{equation}
where $P_{ij}^l$ indicates original image in layer $l$, and $w_l$ are weighting factors compared to the total loss. $A^l$ is the style presentation in layer $l$. Hereby, we get the total loss function: $L_{total} = \alpha L_{content} + \beta L_{style}$, in which $\alpha$ and $\beta$ are weighting factors for content and style construction, respectively. Through times of training, the ratio $\frac{\alpha}{\beta} = \frac{15}{85}$ gave us the best result. Therefore, we adopt this ratio in this method.

\begin{figure}[h!]
  \centering
  \begin{subfigure}[T]{0.32\linewidth}
    \includegraphics[width=\linewidth, height=0.8\linewidth]{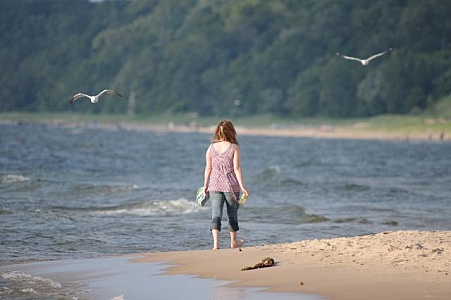}
    \caption{Original Image}
  \end{subfigure}
  \begin{subfigure}[T]{0.32\linewidth}
    \includegraphics[width=\linewidth, height=0.8\linewidth]{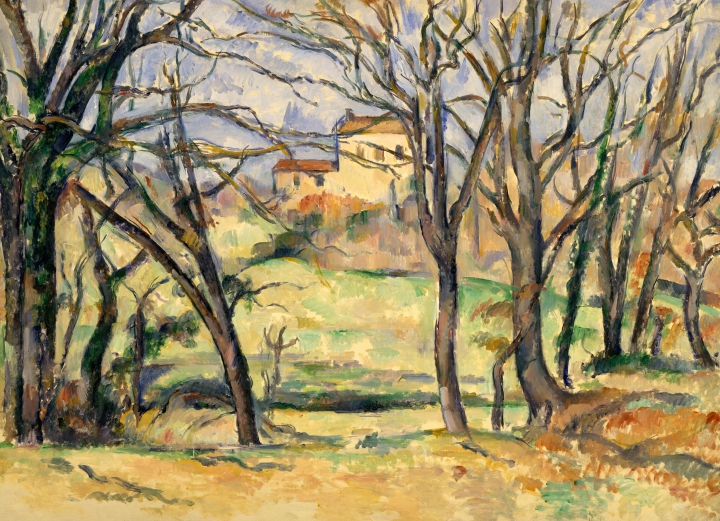}
    \caption{Style Image}
  \end{subfigure}
  \begin{subfigure}[T]{0.32\linewidth}
  \includegraphics[width=\linewidth, height=0.8\linewidth]{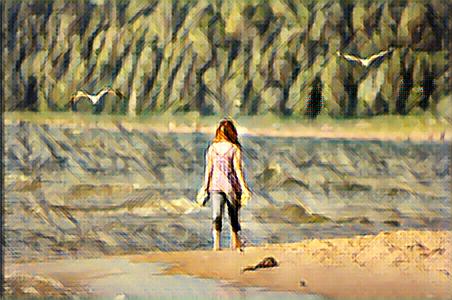}
    \caption{Global Styling}
    \label{fig:global_style_transfer}
  \end{subfigure}
  \begin{subfigure}[T]{0.32\linewidth}
  \includegraphics[width=\linewidth, height=0.8\linewidth]{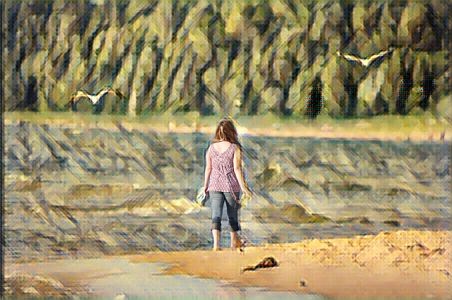}
    \caption{Bg Style Transfer}
    \label{fig:background_style_transfer}
  \end{subfigure}
  \begin{subfigure}[T]{0.32\linewidth}
  \includegraphics[width=\linewidth, height=0.8\linewidth]{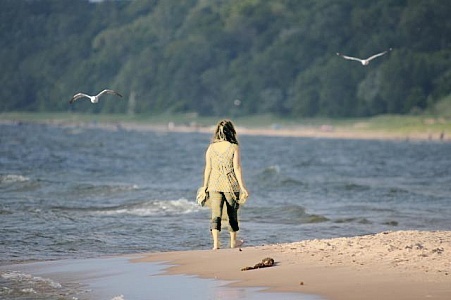}
    \caption{Fg Color Transfer}
    \label{fig:foreground_color_transfer}
  \end{subfigure}
  \begin{subfigure}[T]{0.32\linewidth}
  \includegraphics[width=\linewidth, height=0.8\linewidth]{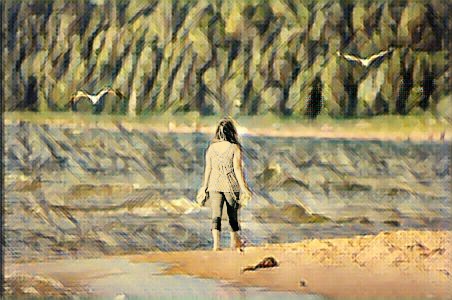}
    \caption{Blend Bg and Fg}
    \label{fig:blend}
  \end{subfigure}
  \caption{showcases the effects of global style transfer (Fig~\ref{fig:global_style_transfer}, background style transfer (Fig~\ref{fig:background_style_transfer}), foreground color transfer (Fig~\ref{fig:foreground_color_transfer}) and final blend image (Fig~\ref{fig:blend}).}
  \label{fig:background_style_transfer_and_foreground_color_transfer}
\end{figure}

\begin{figure*}[h]
  \centering
  \begin{subfigure}[T]{0.15\textwidth}
     \includegraphics[width=1\columnwidth, height=0.7\columnwidth]{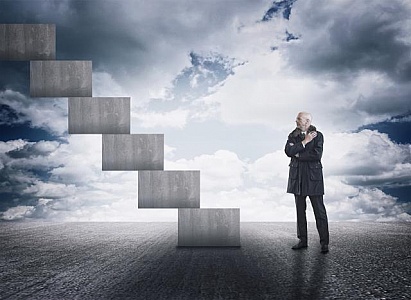}
  \end{subfigure}
  ~
  \begin{subfigure}[T]{0.15\textwidth}
     \includegraphics[width=1\columnwidth, height=0.7\columnwidth]{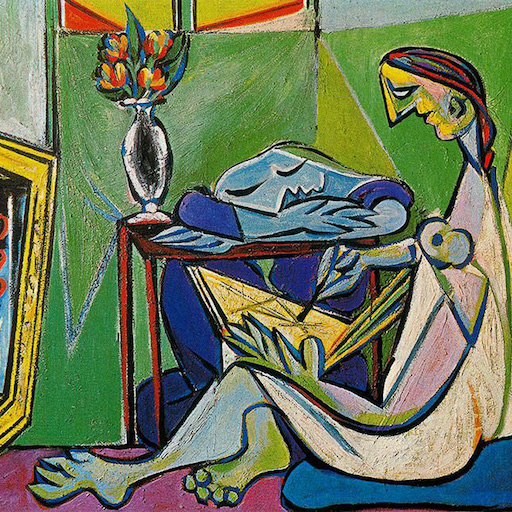}
  \end{subfigure}
  ~
  \begin{subfigure}[T]{0.15\textwidth}
     \includegraphics[width=1\columnwidth, height=0.7\columnwidth]{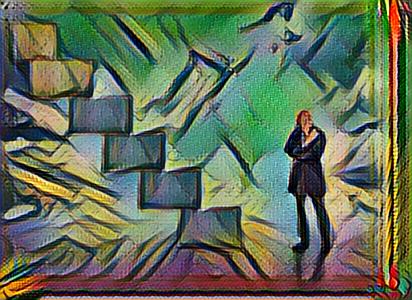}
  \end{subfigure}
  ~
  \begin{subfigure}[T]{0.15\textwidth}
     \includegraphics[width=1\columnwidth, height=0.7\columnwidth]{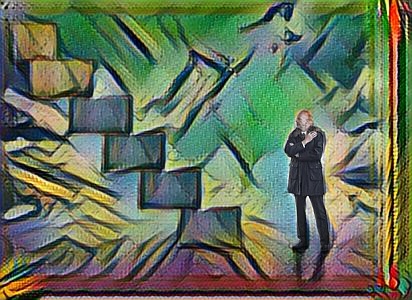}
  \end{subfigure}
  ~
  \begin{subfigure}[T]{0.15\textwidth}
     \includegraphics[width=1\columnwidth, height=0.7\columnwidth]{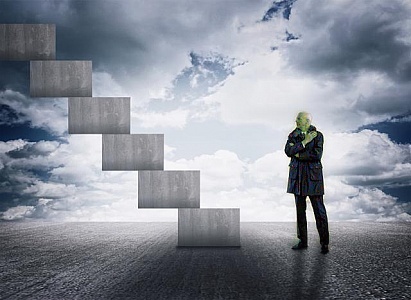}
  \end{subfigure}
  ~
  \begin{subfigure}[T]{0.15\textwidth}
     \includegraphics[width=1\columnwidth, height=0.7\columnwidth]{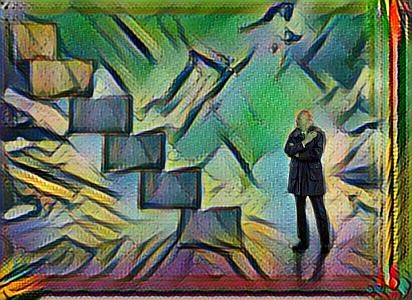}
  \end{subfigure}
  \setcounter{subfigure}{0}
  \hfill
  
  \begin{subfigure}[T]{0.15\textwidth}
     \includegraphics[width=1\columnwidth, height=0.7\columnwidth]{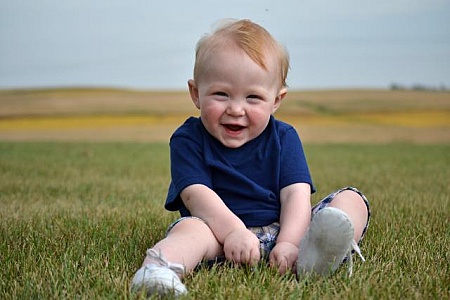}
  \end{subfigure}
  ~
  \begin{subfigure}[T]{0.15\textwidth}
     \includegraphics[width=1\columnwidth, height=0.7\columnwidth]{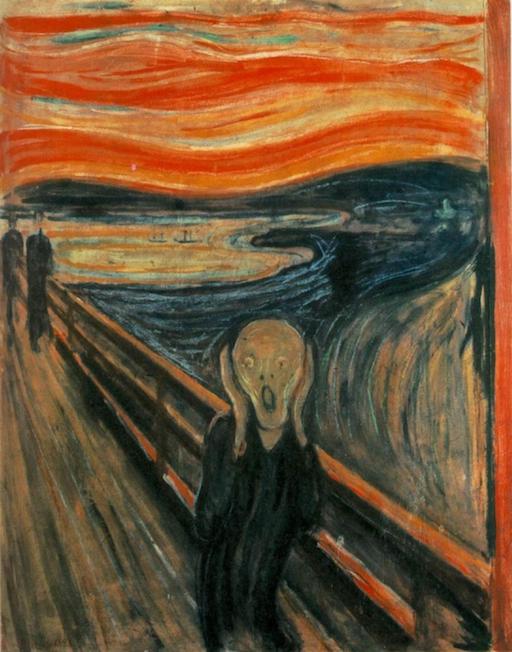}
  \end{subfigure}
  ~
  \begin{subfigure}[T]{0.15\textwidth}
     \includegraphics[width=1\columnwidth, height=0.7\columnwidth]{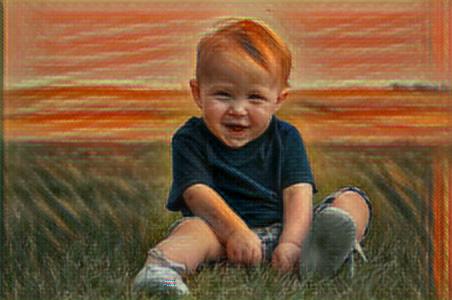}
  \end{subfigure}
  ~
  \begin{subfigure}[T]{0.15\textwidth}
     \includegraphics[width=1\columnwidth, height=0.7\columnwidth]{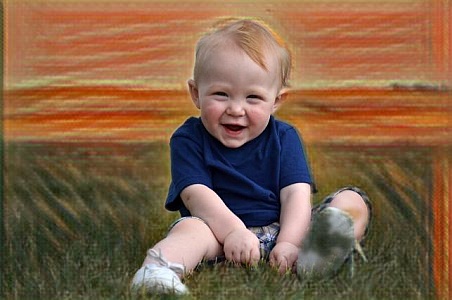}
  \end{subfigure}
  ~
  \begin{subfigure}[T]{0.15\textwidth}
     \includegraphics[width=1\columnwidth, height=0.7\columnwidth]{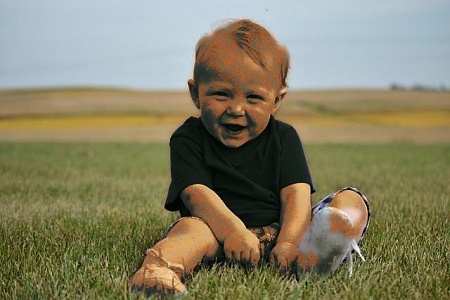}
  \end{subfigure}
  ~
  \begin{subfigure}[T]{0.15\textwidth}
     \includegraphics[width=1\columnwidth, height=0.7\columnwidth]{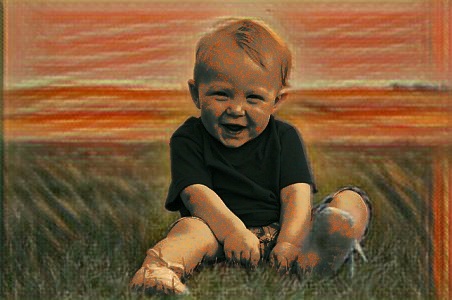}
  \end{subfigure}
  \setcounter{subfigure}{0}
  \hfill

  \begin{subfigure}[T]{0.15\textwidth}
     \includegraphics[width=1\columnwidth, height=0.7\columnwidth]{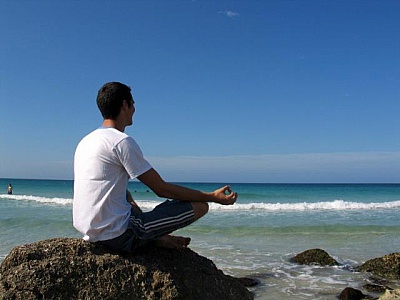}
  \end{subfigure}
  ~
  \begin{subfigure}[T]{0.15\textwidth}
     \includegraphics[width=1\columnwidth, height=0.7\columnwidth]{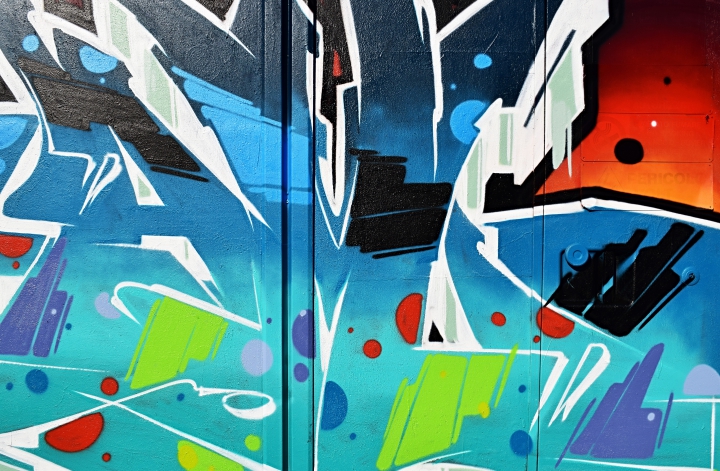}
  \end{subfigure}
  ~
  \begin{subfigure}[T]{0.15\textwidth}
     \includegraphics[width=1\columnwidth, height=0.7\columnwidth]{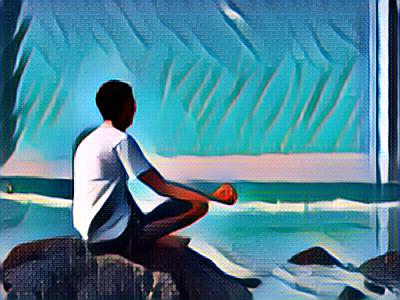}
  \end{subfigure}
  ~
  \begin{subfigure}[T]{0.15\textwidth}
     \includegraphics[width=1\columnwidth, height=0.7\columnwidth]{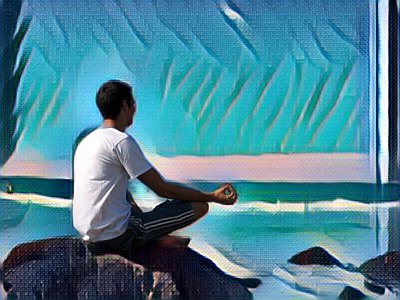}
  \end{subfigure}
  ~
  \begin{subfigure}[T]{0.15\textwidth}
     \includegraphics[width=1\columnwidth, height=0.7\columnwidth]{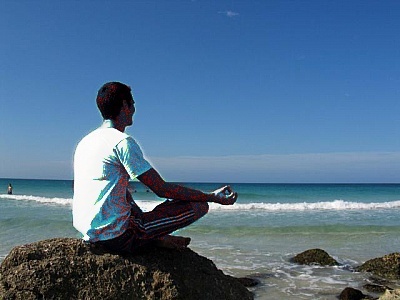}
  \end{subfigure}
  ~
  \begin{subfigure}[T]{0.15\textwidth}
     \includegraphics[width=1\columnwidth, height=0.7\columnwidth]{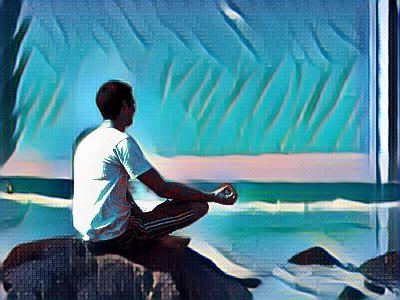}
  \end{subfigure}
  \setcounter{subfigure}{0}
  \hfill
  

  \begin{subfigure}[T]{0.15\textwidth}
     \includegraphics[width=1\columnwidth, height=0.7\columnwidth]{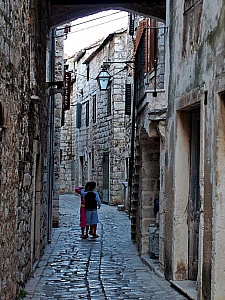}
  \end{subfigure}
  ~
  \begin{subfigure}[T]{0.15\textwidth}
     \includegraphics[width=1\columnwidth, height=0.7\columnwidth]{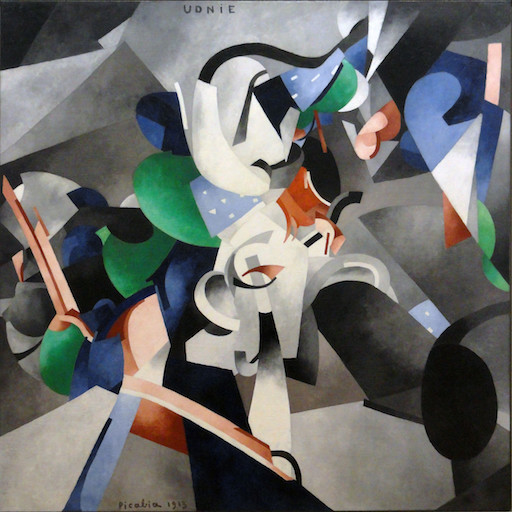}
  \end{subfigure}
  ~
  \begin{subfigure}[T]{0.15\textwidth}
     \includegraphics[width=1\columnwidth, height=0.7\columnwidth]{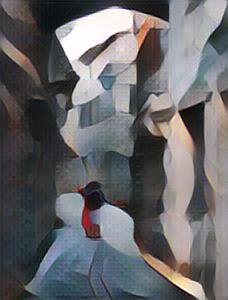}
  \end{subfigure}
  ~
  \begin{subfigure}[T]{0.15\textwidth}
     \includegraphics[width=1\columnwidth, height=0.7\columnwidth]{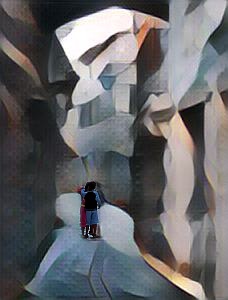}
  \end{subfigure}
  ~
  \begin{subfigure}[T]{0.15\textwidth}
     \includegraphics[width=1\columnwidth, height=0.7\columnwidth]{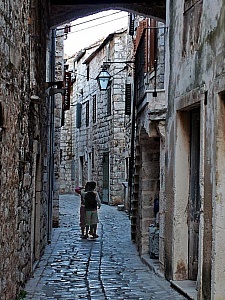}
  \end{subfigure}
  ~
  \begin{subfigure}[T]{0.15\textwidth}
     \includegraphics[width=1\columnwidth, height=0.7\columnwidth]{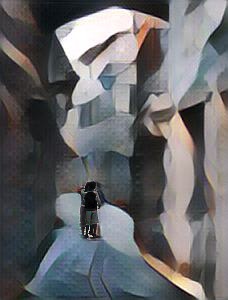}
  \end{subfigure}
  \setcounter{subfigure}{0}
  \hfill
  

  \begin{subfigure}[T]{0.15\textwidth}
     \includegraphics[width=1\columnwidth, height=0.7\columnwidth]{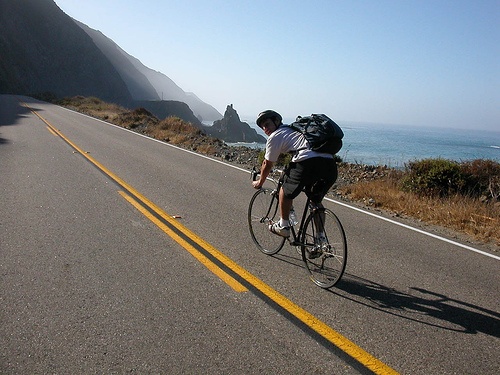}
  \end{subfigure}
  ~
  \begin{subfigure}[T]{0.15\textwidth}
     \includegraphics[width=1\columnwidth, height=0.7\columnwidth]{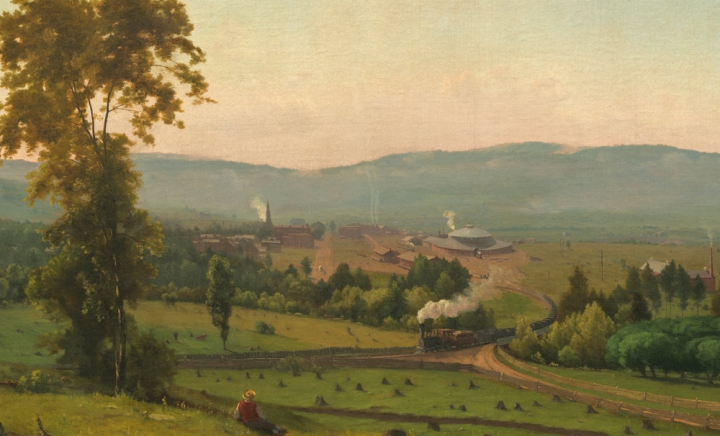}
  \end{subfigure}
  ~
  \begin{subfigure}[T]{0.15\textwidth}
     \includegraphics[width=1\columnwidth, height=0.7\columnwidth]{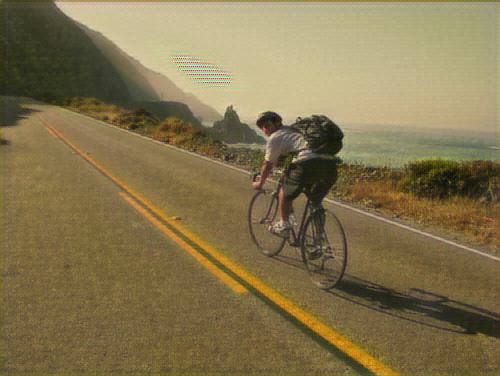}
  \end{subfigure}
  ~
  \begin{subfigure}[T]{0.15\textwidth}
     \includegraphics[width=1\columnwidth, height=0.7\columnwidth]{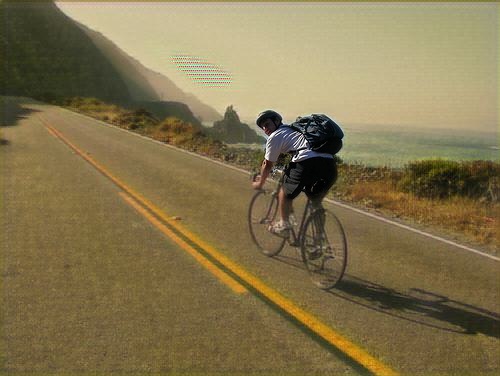}
  \end{subfigure}
  ~
  \begin{subfigure}[T]{0.15\textwidth}
     \includegraphics[width=1\columnwidth, height=0.7\columnwidth]{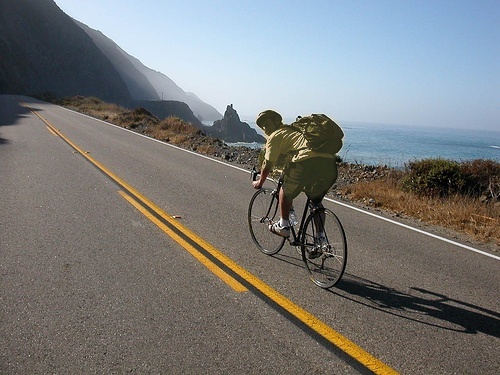}
  \end{subfigure}
  ~
  \begin{subfigure}[T]{0.15\textwidth}
     \includegraphics[width=1\columnwidth, height=0.7\columnwidth]{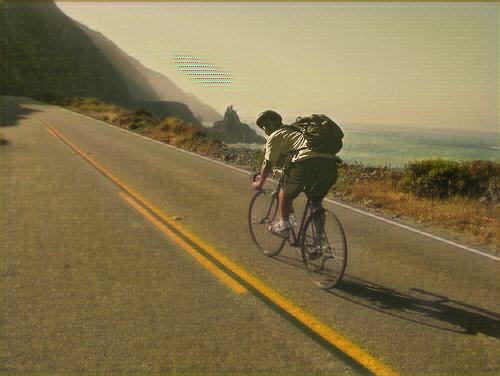}
  \end{subfigure}
  \setcounter{subfigure}{0}
  \hfill
  
  \begin{subfigure}[T]{0.15\textwidth}
     \includegraphics[width=1\columnwidth, height=0.7\columnwidth]{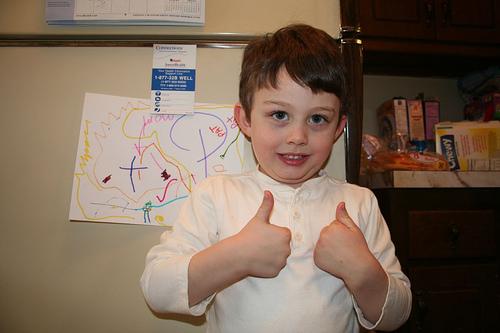}
     \caption{Input Image}
  \end{subfigure}
  ~
  \begin{subfigure}[T]{0.15\textwidth}
     \includegraphics[width=1\columnwidth, height=0.7\columnwidth]{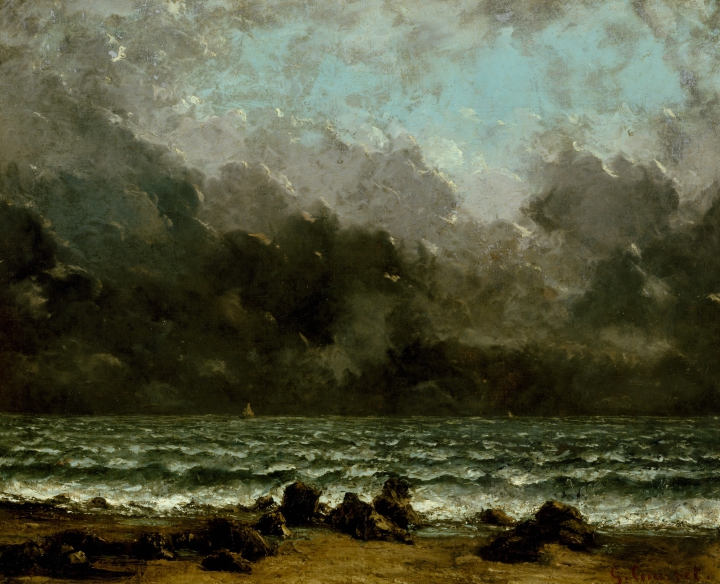}
     \caption{Style Image}
  \end{subfigure}
  ~
  \begin{subfigure}[T]{0.15\textwidth}
     \includegraphics[width=1\columnwidth, height=0.7\columnwidth]{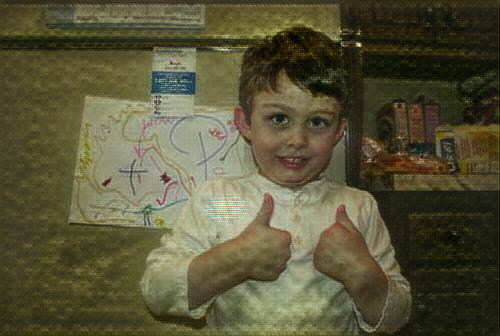}
     \caption{Global Styling}
  \end{subfigure}
  ~
  \begin{subfigure}[T]{0.15\textwidth}
     \includegraphics[width=1\columnwidth, height=0.7\columnwidth]{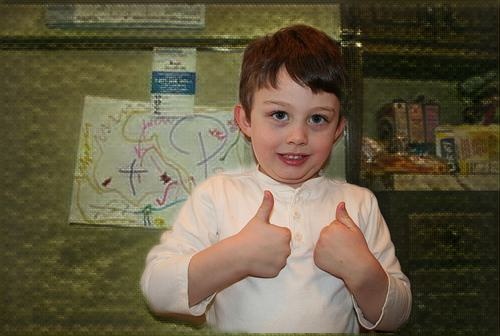}
     \caption{Background-only}
  \end{subfigure}
  ~
  \begin{subfigure}[T]{0.15\textwidth}
     \includegraphics[width=1\columnwidth, height=0.7\columnwidth]{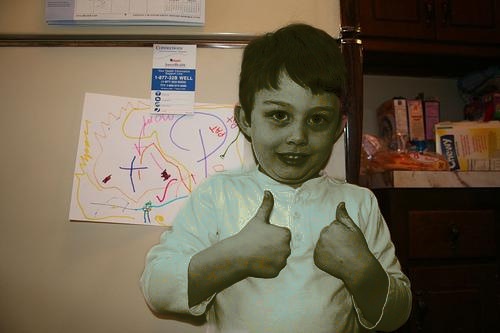}
     \caption{Foreground-only}
  \end{subfigure}
  ~
  \begin{subfigure}[T]{0.15\textwidth}
     \includegraphics[width=1\columnwidth, height=0.7\columnwidth]{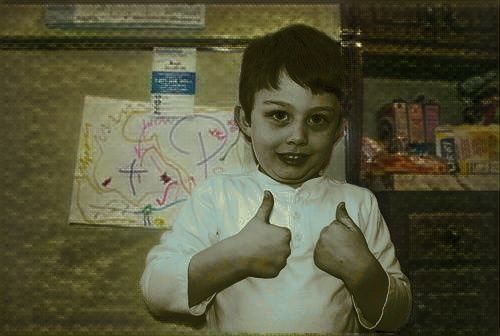}
     \caption{\textbf{Ours}}
  \end{subfigure}
  \caption{illustrates a qualitative comparison of the results obtained for six sets of input image (a) and their corresponding style image (b). The comparison includes four images manipulation techniques: global style transfer (c), background-only style transfer (d), foreground-only color transfer (e), and our proposed method – regional style and color transfer (f), which combines background style transfer with foreground color transfer.}
  \label{fig:blending-style-and-color}
\end{figure*}

\subsection{Color Transfer}

This work proposes a method for color transfer specifically targeted at the foreground region of an image. This approach aims to achieve color harmony between the foreground (e.g., a person) and a style-transferred background while preserving the integrity of the portrait. To achieve this, we leverage the $l\alpha \beta$ color space, which aligns well with the statistics of natural scenes and human visual perception, for color representation. The conversion from the original RGB space to $l\alpha \beta$ follows the established method by Reinhard \etal \cite{Reinhard}:
\begin{equation} \begin{bmatrix}L\\ M\\ S\\ \end{bmatrix}
=\begin{bmatrix}0.3811 & 0.5783 & 0.0402 \\
0.1967 & 0.7244 & 0.0782 \\
0.0241 & 0.1288 & 0.8444 \end{bmatrix}
\begin{bmatrix}R\\ G\\ B\\ \end{bmatrix}
\end{equation}
\begin{equation} \begin{bmatrix}l\\ \alpha\\ \beta\\ \end{bmatrix}
=\begin{bmatrix}\frac{1}{\sqrt{3}} & 0 & 0 \\
0 & \frac{1}{\sqrt{6}} & 0 \\
0 & 0 & \frac{1}{\sqrt{2}} \end{bmatrix}
\begin{bmatrix}1 & 1 & 1 \\
1 & 1 & -2 \\
1 & -1 & 0 \end{bmatrix}
\begin{bmatrix}\log L\\ \log M\\ \log S\\ \end{bmatrix}
\end{equation}

Within the $l\alpha \beta$ space, each pixel is represented as a three-dimensional vector. To reduce dimension and capture the dominant color variations, we employ Principal Component Analysis (PCA) in this 3D space for both the image and the style representation. This involves eigenvalue decomposition of the covariance matrix $X^TX$, where $X$ is an $n\times 3$ matrix containing all pixel vectors of an image in $l\alpha \beta$ space. The first principal component vector, q, corresponding to the largest eigenvalue, captures the most significant color variation direction. By calculating the dot product with $q$, all pixel vectors in an image are projected onto a one-dimensional line, effectively capturing the main color distribution.

For each pixel $i$ on the original image, our task is to find the equivalent pixel $j$ on the style image and transfer its RGB color:
\begin{equation}
    (R_{T'i},G_{T'i},B_{T'i})\gets(R_{Sj},G_{Sj},B_{Sj})
\end{equation}

\noindent We use the pixel $j$ such that
\begin{equation}
    F[(l_{Ti},\alpha_{Ti},\beta_{Ti})\cdot q_T]=F[(l_{Sj},\alpha_{Sj},\beta_{Sj})\cdot q_S]
\end{equation}
in which $F(x)$ indicates the cumulative density function and $q_T$ and $q_S$ indicate the first principal components of the original image and the style, respectively. 




\subsection{Blending Style and Color} 

In the preceding stages, a background style-transferred image, a foreground color-transferred image, and an optimized person segmentation mask were generated. To create the final result, these elements are seamlessly combined leveraging the optimized segmentation mask.

We employ alpha blending, a well-established technique in computer graphics, to achieve a seamless integration of the aforementioned images. As detailed in Section~\ref{sec:segmentation}, the Euclidean distance metric was previously utilized to generate a feathered boundary within the segmentation mask. This inherent property of the mask provides a range of values between $0.0$ and $1.0$, perfectly suited for alpha blending. The following equation mathematically represents the pixel-wise blending process:

\begin{equation}
C_{blend} = \alpha \cdot C_{color} + (1-\alpha) \cdot C_{style}
\end{equation}

\noindent where $\alpha$ is the pixel value of boundary, $C_{style}$ denotes the pixel value of global style transfer image, and $C_{color}$ represents the pixel value of global color transfer image.

\section{Results}

\subsection{Segmentation and Boundary Optimization}
We evaluated the impact of boundary optimization on regional style transfer. To isolate the effect of boundary optimization, color transfer to the foreground region was intentionally omitted from the analysis. As illustrated in Fig~\ref{segmentation_mask}, the proposed boundary optimization algorithm demonstrably improved boundary preservation, highlighting its effectiveness.

\subsection{Background Style Transfer}
In the context of neural style transfer, we employ a loss function that balances the contributions of the content image and the style reference image. This optimization process leads to a trade-off parameter that determines the relative influence of each image on the final output. Leveraging the segmentation mask obtained in the preceding step, we achieve selective style transfer by applying the artistic style solely to the background region, while preserving the foreground content of the original image. As illustrated in Fig~\ref{fig:background_style_transfer}, the resulting image successfully integrates the content of the original image with the style of the reference image, applied exclusively to the background.

\subsection{Foreground Color Transfer}
To achieve visual harmony between the style-transferred background and the foreground object (e.g., person), we perform a color transfer operation specifically on the foreground region. This process aims to adapt the color palette of the foreground to better complement the stylized background, while maintaining the integrity of the portrait itself. Fig~\ref{fig:foreground_color_transfer} demonstrates the successful integration of the original image content with the color characteristics of the style image, applied selectively to the foreground element.


\subsection{Image Blending}
Leveraging the strengths of both style transfer and color transfer, our approach meticulously separates the foreground and background regions. This enables the creation of a final output where the artistic style from the reference image is seamlessly integrated into the background, while the foreground portraits retain their original color characteristics with rigorous fidelity.  Fig~\ref{fig:blend} exemplifies this meticulous approach, showcasing the generation of an aesthetically unified image. The foreground elements appear to harmoniously co-exist with the artistically rendered background, achieving a visually compelling and perceptually convincing outcome.

To comprehensively evaluate the effectiveness of our proposed method, we conducted a series of experiments utilizing diverse input and style images. Fig~\ref{fig:blending-style-and-color} showcases a comparative analysis of various image manipulation techniques. As can be observed, global style transfer exhibits a significant limitation: the style is indiscriminately applied across the entire image, resulting in a visually distorted and unnatural appearance for both the foreground and background. Background-only style transfer, while successful in preserving the integrity of the foreground portraits, introduces a visual inconsistency, where the portraits appear in-congruent with the artistically transformed background. In contrast, our proposed regional style and color transfer method stands out as a groundbreaking advancement. It achieves the remarkable feat of seamlessly integrating the artistic style from the reference image  into the background, while meticulously preserving the natural color fidelity of the foreground portraits. This results in a visually harmonious image where the foreground elements organically co-exist with the artistically rendered background, creating an aesthetically pleasing and perceptually convincing outcome.

\section{Conclusions}

This paper presents a novel approach to regional style and color transfer, effectively addressing the limitations of global style transfer methods that often produce unnatural results. Our proposed technique leverages the power of regional style and color transfer within selective image regions, demonstrably leading to better visual quality and enhanced realism.

Extensive experimentation underscores the effectiveness of our method in controlling style and color transfer across various regions, culminating in a significantly more natural aesthetic.  The elegance and efficiency of our approach lie in its seamless integration with existing segmentation techniques, enabling style and color transfer to designated image areas.

Furthermore, our framework exhibits inherent scalability, readily adaptable to establish stylistic correspondences between painted images and video frames, paving the way for the creation of stylized films.  While our current implementation demonstrates minimal computational overhead, future research directions include extending the framework's applicability beyond human segmentation to encompass diverse object categories (e.g., vehicles, animals). Additionally, we aim to optimize inference speed and efficiency to address the exponential growth in computational cost associated with larger input images. These advancements will constitute the cornerstone of our ongoing research efforts.

\renewcommand{\bibfont}{\footnotesize}

\footnotesize{
\bibliographystyle{IEEEtran}
\bibliography{main}
}

\end{document}